\def\year{2022}\relax
\documentclass[letterpaper]{article} % DO NOT CHANGE THIS
\usepackage{aaai22}  % DO NOT CHANGE THIS
\usepackage{times}  % DO NOT CHANGE THIS
\usepackage{helvet}  % DO NOT CHANGE THIS
\usepackage{courier}  % DO NOT CHANGE THIS
\usepackage[hyphens]{url}  % DO NOT CHANGE THIS
\usepackage{graphicx} % DO NOT CHANGE THIS
\usepackage{natbib}  % DO NOT CHANGE THIS AND DO NOT ADD ANY OPTIONS TO IT
\usepackage{caption} % DO NOT CHANGE THIS AND DO NOT ADD ANY OPTIONS TO IT
\DeclareCaptionStyle{ruled}{labelfont=normalfont,labelsep=colon,strut=off} % DO NOT CHANGE THIS
\usepackage{algorithm}
\usepackage{algorithmic}
\usepackage{authblk}

\usepackage{tikz}

\usepackage{newfloat}
\usepackage{listings}
\floatstyle{ruled}
\newfloat{listing}{tb}{lst}{}
\floatname{listing}{Listing}
\title{ARDIAS: AI-Enhanced Research Management, Discovery, and Advisory System}

\author[1]{Debayan Banerjee}
\author[2]{Seid Muhie Yimam}
\author[1]{Sushil Awale}
\author[1,2]{Chris Biemann}
\affil[1]{Language Technology Group, Department of Informatics, Universität Hamburg, Germany}
\affil[2]{House of Computing and Data Science, Universität Hamburg, Germany}
\affil[ ]{\textit {\{debayan.banerjee,seid.muhie.yimam,sushil.awale,christian.biemann\}@uni-hamburg.de}}

\usepackage{bibentry}
\begin{document}

\maketitle

\begin{abstract}

In this work, we present ARDIAS, a web-based application that
aims to provide researchers with a full suite of discovery and collaboration tools. ARDIAS currently allows searching for authors and articles by name and gaining insights into the research topics of a particular researcher. With the aid of AI-based tools, ARDIAS aims to recommend potential collaborators and topics to researchers. In the near future, we aim to add tools that allow researchers to communicate with each other and start new projects.

\end{abstract}

% contents from the ARDIAS Proposal -- should be adapted in a paper format
\section{Introduction}

Historically, scientific research has been performed in closely-knit groups inside individual institutes or corporations, with little to no collaboration across different geographies \cite{Gary2008}. The usual mode of exchange of scientific findings used to be publishing articles in journals, and more rarely,  researchers used to meet each other in person during scientific conferences.  

With the advent of the Internet, it is finally possible to follow current research topics and researchers more closely online. Institutions host on their websites a collection of publications by their researchers,  along with some contact details. Moreover, conferences and journals often publish their proceedings online for the public to access. In the case of open-access publications, people may access the documents free of cost, however, for paid publications, users must either pay a fee or belong to an affiliated institute to access these articles.

To make such research even more accessible, in the past decade, a number of aggregators of research articles have appeared online. Some notable products and projects in this area are DBLP\footnote{\url{https://dblp.org/}}, OpenAlex\footnote{\url{https://openalex.org/}}, Research Gate\footnote{\url{https://www.researchgate.net/}}, Microsoft Academic Network, Semantic Scholar\footnote{\url{https://www.semanticscholar.org/}}, Google Scholar\footnote{\url{https://scholar.google.com/}} and ORKG\footnote{\url{https://orkg.org/}}. 

In spite of the existing tools available for the discovery of research work, a major gap remains in tools that allow collaboration. ARDIAS aims to enable the following workflow:

\begin{itemize}
  \item User visits ARDIAS (either logged in or as a guest)
  \item User discovers relevant research and authors
  \item User communicates with the relevant author of interest and starts a new research project on the platform
  \item The project members track experimental progress and log results on the platform
  \item The research article is drafted on ARDIAS and submitted to a conference or journal
\end{itemize}

\begin{figure*}[h]
\centering
    \includegraphics[width=0.85\linewidth]{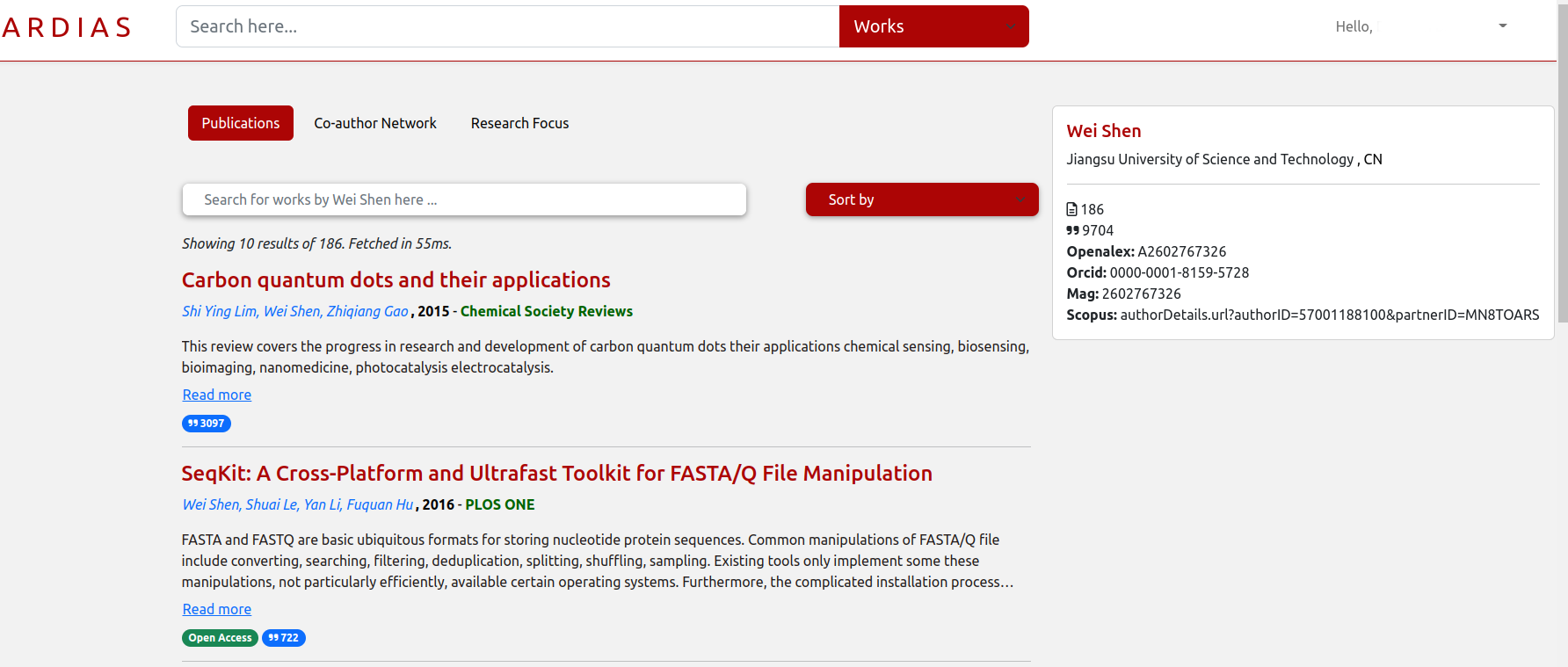}
\caption{Author search results in ARDIAS.}
\label{fig:author}
\end{figure*}

\section{Related Work}

For the purpose of scholarly communication and its methods, most solutions so far have emphasized on techniques for
linking metadata about articles, people, data and other relevant concepts \cite{10.4225/03/58c696655af8ab, scholix}. The common mode today is to extend the representation to a document structure \cite{alma991016811776206926}. Some other solutions propose more comprehensive conceptual models for scholarly knowledge that capture problems, methods, theories, statements, concepts, and their relations \cite{723935, Meister2017TowardsAK, 10.1145/3360901.3364435}. Our focus is not to reinvent any of these methods, but to use them as a baseline system for our initial discovery phase within ARDIAS.

The next step in ARDIAS is a recommendation component, which has seen significant activity \cite{ISINKAYE2015261} in machine learning research in general, and also in the focused area of scholarly research \cite{Xia_2016, F_rber_2020}. For the purpose of recommending relevant future research topics, potential co-authors, and relevant citations, one of the required steps is to extract relevant information from a scientific manuscript and then process it. In this regard, there are plenty of works, particularly in the area of information extraction from scholarly documents  \cite{10.1007/s11192-018-2921-5}. 

The task of finding relevant people with a given expertise is called Expert Finding, and several works have focused on this problem \cite{10.1145/1148170.1148181, 10.1007/978-3-540-71496-5_38, 10.1145/1458082.1458232, fischer-etal-2019-lt}. For this task certain datasets have been developed, for example, the enriched version of DBLP \footnote{\url{https://aminer.org/lab-datasets/expertfinding/}} provided by the ArnetMiner project \cite{10.1145/1401890.1402008}, or the W3C Corpus \footnote{\url{https://tides.umiacs.umd.edu/webtrec/trecent/parsed_w3c_corpus.html}} of TREC used by \citet{10.1145/1148170.1148181}.

From a communication perspective, the internet has provided several new avenues for interaction among researchers. It has been shown that social media is an effective means of collaborative learning in new domains \cite{Ansari2020}. Researchers are active on privately owned platforms like Linkedin and Twitter, however, very recently there has been a migration of researchers from Twitter to Mastodon\footnote{\url{https://mastodon.social/explore}}. Mastodon is an open-source publish-subscribe protocol-based broadcast platform that is distributed and self-hosted in nature. Among the current top Mastodon servers for the specific category of AI researchers, \emph{sigmoid.social} seems to be the most popular. Hosting a Mastodon server, and an XMPP server to enable real-time collaboration is a possible future direction that ARDIAS may take to enable communication among researchers.

For collaboration, several products and projects exist that allow tracking, collection, analysis, and visualization of scientific results. Some such platforms include Overleaf, TensorBoard, Wandb, Trello, GitHub, and others. In ARDIAS, it is our endeavor to select the best features from existing products and implement an open-source version for them.

\section{Architecture of the System}
%\subsection{Approaches}

\begin{figure}[h]
\includegraphics[width=0.35\textheight]{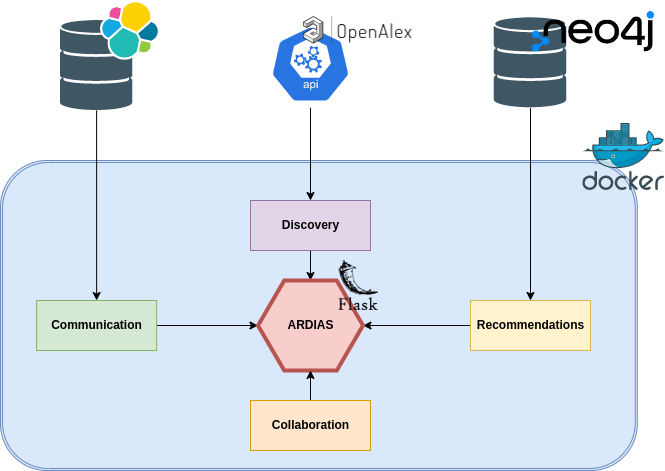}
\caption{ARDIAS architecture}
\label{fig:arch}
\end{figure}

Figure \ref{fig:arch} shows the different components of ARDIAS. In the back end, ARDIAS currently depends on the OpenAlex API to fetch and update the latest research articles and corresponding metadata. A local copy of the OpenAlex data is stored in an Elasticsearch instance. Additionally, the OpenAlex data dump is converted into a Knowledge Graph (KG) and indexed into a graph database hosted in Neo4J locally. The front end currently interfaces with 1) OpenAlex API 2) Elasticsearch and 3) Neo4J to display relevant information to the user. A batch of helper scripts downloads the latest OpenAlex updates and updates the local Neo4J graph. 

At the moment we do not have a local database for storing user data, settings, or preferences. Hence the user experience is largely built for a user who has not registered or logged in. However, a login feature does exist that stores login information in memory per session. In the near future, we plan to implement a more comprehensive login and registration system which verifies users thoroughly before onboarding them to ARDIAS.

In the future, additional infrastructure for machine learning and big data ingestion shall be added to ARDIAS.

\begin{figure}[h]
\includegraphics[width=0.35\textheight]{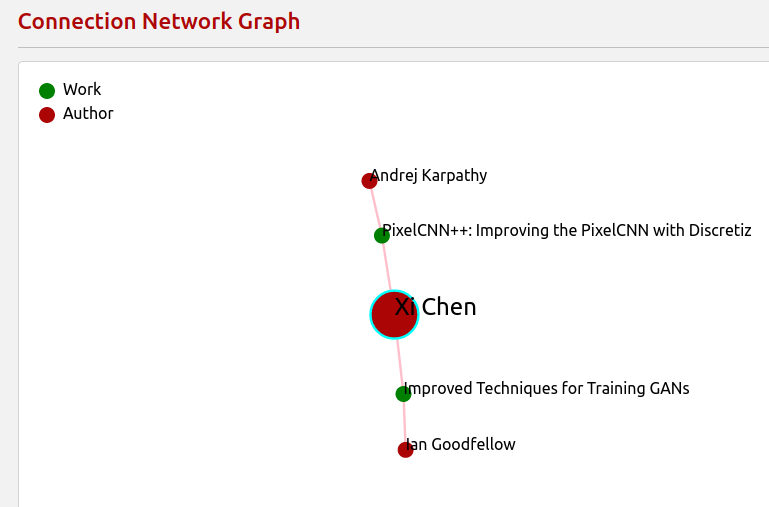}
\caption{Shortest authorship path between two authors}
\label{fig:shortpath}
\end{figure}

\begin{figure}[h]
\includegraphics[width=0.35\textheight]{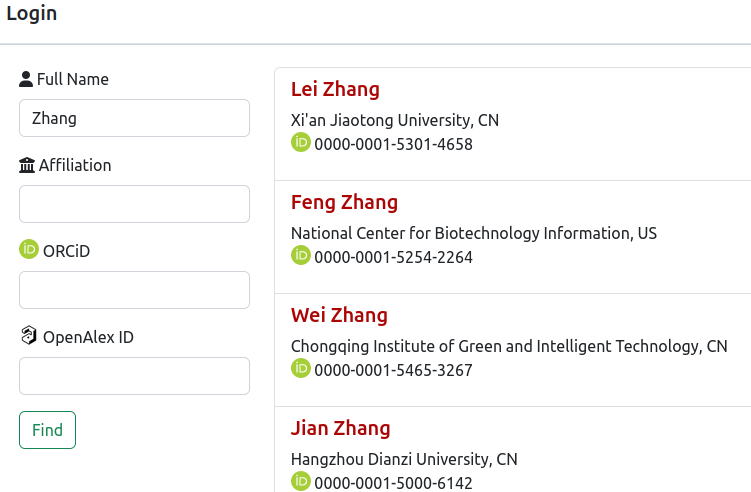}
\caption{Author disambiguation for first-time login.}
\label{fig:login}
\end{figure}

\section{Features of ARDIAS}
\subsection{Discovery}
The first stage of the workflow of ARDIAS is discovery. This begins with user login, where a first-time user is presented with the view shown in Figure \ref{fig:login}. The OpenAlex data we consume may contain several authors of the same name and several orphan nodes for the same author. Hence, it is important to ask the authors to identify themselves on the first login, and to solve this problem, our login page produces a list of authors by text match over name and affiliation. The user can then select the correct author from the list and self-identify. Additionally, it is also possible for a user to explore most features of ARDIAS without logging in. 
Scholarly work discovery begins with the discovery of one's own work on the platform. If the user is logged in, ARDIAS fetches the works via an OpenAlex API call and populates the Works page. If the user is not logged in, the user can search for himself or herself using the prominently located search bar in the UI. The search box allows a user to fetch relevant results based on a handful of different criteria, namely 1) Works 2) Author 3) Institutions 4) Venues. 
\subsubsection{Works} Searching by works criterion displays a paginated results page showing the relevant publications based on a text search of the title.  The page contains the title of the paper, the list of authors, the publication year, and the venue. A short abstract follows, and at the bottom, the citation count is displayed. The author names and venue names are hyperlinks that may be clicked, and they open to the respective author and venue pages. On clicking the title of the work itself, a dedicated publication page is loaded which additionally displays DOI information, OpenAlex and Microsoft Academic Graph (MAG) IDs, and keywords for this work. A tab is also available to view the citations of this work from where the citing papers may also be visited. The next section of the page displays similar papers to the current work. The last section of the Work page contains a discussion section, where logged-in users may add comments and start a discussion on the paper topic.
\subsubsection{Authors}As depicted in Figure \ref{fig:author}, searching by author fetches a list of authors matched by text search, with a short display of their institute of affiliation, the number of publications, citation counts, and OpenAlex and MAG IDs. Several authors may share the same name, hence the institution name is helpful in disambiguating the authors. On clicking on an author's name on this page, a more detailed author page opens, which displays the published works of the author in sorted order of citation counts. If a publication in this list is Open Access, it is marked as so. The results on this page may be further sorted based on title, date, and citations. In a separate tab, one can see the co-author network for the author, arranged in a hot-spot-styled graphic. The subsequent tab displays the research focus areas of the author in a similar style. 
\subsubsection{Institution}Searching by institution displays a paginated list of institutes based on text search similarity of the institute name. The title for each institute displays the name and location, while the subsequent lines display the website address, the sector of establishment, and the acronym popularly used for the institute if any. This short selection of information allows disambiguation of similarly named institutes. On clicking on an institute, a dedicated institute page opens, which on the right side displays the institute logo, institute name, location, homepage, Wikipedia links, publication, and citation counts. In the central section of the page, a list of authors belonging to this institute sorted by citation counts is displayed.
\subsubsection{Venue}Searching by venue displays a list of similarly named venues with corresponding historical publication and citation counts. Clicking on a venue opens a new page that lists the publications for the given venue in sorted order based on citations. This sorting order can be changed to be based on title and date. 

\subsection{Recommendation}

Users of ARDIAS also receive recommendations from the system on a number of different categories, both with and without the aid of Machine Learning methods.\\
Some basic non-ML recommendations are as follows: 1) ARDIAS produces a list of researchers who work at the same institute as the user 2) ARDIAS is able to show the least number of hops between a user and another researcher on the platform through work and author nodes. 3) ARDIAS produces a list of related sub-topics of interest that are currently popular in the research community.\\
Some ML-based recommendations we plan for ARDIAS in the future include 1) a list of similar researchers based on common interests. This would be achieved via graph-based node embedding algorithms and nearest neighbor computations of the embedding space 2) a list of similar works based on common publications using similar node embedding methods 3) a list of potential and high-impact research topics based on node prediction approaches in graph-based ML.\\
Apart from recommendations in the discovery process of research and authors, we see scope for recommendations algorithms in the manuscript authoring phase as well. For example, a beginner scholar may be making elementary formatting, grammar, spelling, or structural errors in the paper being authored. In some cases, a significant section like Related Work may be missing. In such cases, AI-based technologies will be able to help the scholar with helpful suggestions on how to improve the draft.\\
Another area where AI-based recommendations may help in the authoring process is the automatic suggestion of citations for the given topic of the manuscript, providing an easy selection between different formats of citations within the Latex editor window, eg: BibTex, Crossref, etc.\\
A special aspect of non-ML recommendation is depicted in Figure \ref{fig:shortpath}. If author A wishes to work with author B, ARDIAS fetches the shortest path between the two nodes in the scholarly KG hosted in the graph database. This allows author A to find common connections to author B. This makes communication and collaboration stages possible for author A.

\subsection{Communication}

After the phases of discovery and recommendation, an author may want to get in touch with another author on our platform. At the moment we have not implemented any communication method explicitly, however, the vision is to enable this kind of communication through an open-source and well-established messaging platform. Specific methods need to be developed to keep spam at bay, and verification of users on our platform remains vital. Some methods for logged-in user verification may include making an institution or work e-mail mandatory for sign-up, and also having to supply an ORCID or a similar academic network ID. \\
Recently there has been an exodus of academics shifting from the social media site of Twitter to Mastodon. Twitter is a privately held social media company that allows micro-blogging, while Mastodon is a free and open-source software that allows the hosting of independent and federated nodes, each with its own set of policies. ARDIAS may host a dedicated Mastodon node for researchers to join and also allow communication across other Mastodon nodes on our platform. From a communications perspective, we additionally plan to explore the XMPP protocol to allow chat between researchers who already accept each other's request to communicate.

\subsection{Collaboration}

After researchers have discovered each other and decided to collaborate on a project, ARDIAS aims to provide a basic platform to start and manage a project. At the moment the components are still in the planning phase, however, some basic requirements for project collaboration are clear. On the creation of a project on the platform, the collaborating researchers should gain access to an online document editing and storage platform, which allows not just plain text documents, but also binary files, models, and archived logs. The project space should have an issue tracker so planning on individual components of the project is possible and progress is trackable. It would also be desirable to provide an online platform that allows various forms of visualization of results, and logging of experimental data. Finally, it would be desirable to allow editing of the research manuscript itself on the platform, perhaps through a LaTeX editor.

\section{Use cases and applications}
  \label{usecases}
\noindent \textbf{Searching for research projects and collaborators}: ARDIAS shall store information about users and their expertise, lists of tools and datasets, machine learning models and their applications, detailed documentation of projects, concept notes, as well as research publications. Traditionally, one should look for a research collaborator either by searching online profiles, seeking or contacting the responsible office in the research institute, or advertising a call for collaboration. ARDIAS can help by automatically searching or recommending collaborators using the knowledge stored in the repository. The system ranks and recommends collaborators based on their expertise and relevance. \\
\noindent \textbf{Research approach suggestion}: Research on a multidisciplinary project is challenging. Medical researchers are aware of problems in their discipline but they might lack the proper approaches to build an automated system with a machine learning component. Similarly, in a computational linguistic research project, experts from different fields such as linguistics, mathematics, statistics, and computer science work together. Employing computational linguistics approaches for social science projects needs a further understanding of different research problems.  Using an AI-enhanced system, where prior multidisciplinary projects are indexed, can help in suggesting research approaches for new research problems. \\
%\noindent \textbf{Profile matching}: A knowledge management (KM) system that stores experts and their profiles will have the potential for automatic profile matching. If the system is fed with the research description, it will match the expert who can fit the problem description. The KM system can be used, for example, to assign a research proposal or paper to an appropriate reviewer.\\
\noindent \textbf{Virtual help-desk}: In large research institutes or companies, where there are several research projects, tools, and research collaborators, it is difficult to provide accurate information. A research knowledge management system with interactive UI can help in finding requested services that can be enhanced with faceted search functionality to support help desk professionals.

\section{Conclusion}

In this work, we described ARDIAS, which is a web application with a vision for researchers to discover, communicate and collaborate on new research topics. The demo is made available via a public URL\footnote{\url{https://ardias.ltdemos.informatik.uni-hamburg.de/}}. The development is currently at an initial phase, with basic features having been implemented. We present our long-term goal for ARDIAS as being a common platform for researchers from all backgrounds and fields to discover each other and collaborate. 

\section{Acknowledgements}

ARDIAS is funded by the "Idea and Venture Fund" research grant by Universität Hamburg, which is part of the Excellence Strategy of the Federal and State Governments.
\bibliography{aaai22}

\begin{thebibliography}{17}
\providecommand{\natexlab}[1]{#1}

\bibitem[{Ansari and Khan(2020)}]{Ansari2020}
Ansari, J. A.~N.; and Khan, N.~A. 2020.
\newblock Exploring the role of social media in collaborative learning the new
  domain of learning.
\newblock \emph{Smart Learning Environments}, 7(1): 9.

\bibitem[{Aryani and Wang(2017)}]{10.4225/03/58c696655af8ab}
Aryani, A.; and Wang, J. 2017.
\newblock {Research Graph: Building a Distributed Graph of Scholarly Works
  using Research Data Switchboard}.

\bibitem[{Balog, Azzopardi, and de~Rijke(2006)}]{10.1145/1148170.1148181}
Balog, K.; Azzopardi, L.; and de~Rijke, M. 2006.
\newblock Formal Models for Expert Finding in Enterprise Corpora.
\newblock In \emph{Proceedings of the 29th Annual International ACM SIGIR
  Conference on Research and Development in Information Retrieval}, SIGIR '06,
  43–50. New York, NY, USA: Association for Computing Machinery.
\newblock ISBN 1595933697.

\bibitem[{{Brodaric, Boyan and Reitsma, Femke and Qiang, Yi}({2008})}]{723935}
{Brodaric, Boyan and Reitsma, Femke and Qiang, Yi}. {2008}.
\newblock {SKIing with DOLCE : toward an e-Science knowledge infrastructure}.
\newblock In {Eschenbach, C and Gruninger, M}, ed., \emph{{Frontiers in
  Artificial Intelligence and Applications}}, volume {183}, {208--219}.

\bibitem[{Burton et~al.(2017)Burton, Koers, Manghi, Stocker, Fenner, Aryani,
  La~Bruzzo, Diepenbroek, and Schindler}]{scholix}
Burton, A.; Koers, H.; Manghi, P.; Stocker, M.; Fenner, M.; Aryani, A.;
  La~Bruzzo, S.; Diepenbroek, M.; and Schindler, U. 2017.
\newblock The Scholix Framework for Interoperability in Data-Literature
  Information Exchange.
\newblock \emph{D-Lib Magazine}, 23.

\bibitem[{Fang and Zhai(2007)}]{10.1007/978-3-540-71496-5_38}
Fang, H.; and Zhai, C. 2007.
\newblock Probabilistic Models for Expert Finding.
\newblock In Amati, G.; Carpineto, C.; and Romano, G., eds., \emph{Advances in
  Information Retrieval}, 418--430. Berlin, Heidelberg: Springer Berlin
  Heidelberg.
\newblock ISBN 978-3-540-71496-5.

\bibitem[{Fischer, Remus, and Biemann(2019)}]{fischer-etal-2019-lt}
Fischer, T.; Remus, S.; and Biemann, C. 2019.
\newblock {LT} Expertfinder: An Evaluation Framework for Expert Finding
  Methods.
\newblock In \emph{Proceedings of the 2019 Conference of the North {A}merican
  Chapter of the Association for Computational Linguistics (Demonstrations)},
  98--104. Minneapolis, Minnesota: Association for Computational Linguistics.

\bibitem[{Färber and Jatowt(2020)}]{F_rber_2020}
Färber, M.; and Jatowt, A. 2020.
\newblock Citation recommendation: approaches and datasets.
\newblock \emph{International Journal on Digital Libraries}, 21(4): 375--405.

\bibitem[{Isinkaye, Folajimi, and Ojokoh(2015)}]{ISINKAYE2015261}
Isinkaye, F.; Folajimi, Y.; and Ojokoh, B. 2015.
\newblock Recommendation systems: Principles, methods and evaluation.
\newblock \emph{Egyptian Informatics Journal}, 16(3): 261--273.

\bibitem[{Jaradeh et~al.(2019)Jaradeh, Oelen, Farfar, Prinz, D'Souza,
  Kismih\'{o}k, Stocker, and Auer}]{10.1145/3360901.3364435}
Jaradeh, M.~Y.; Oelen, A.; Farfar, K.~E.; Prinz, M.; D'Souza, J.; Kismih\'{o}k,
  G.; Stocker, M.; and Auer, S. 2019.
\newblock Open Research Knowledge Graph: Next Generation Infrastructure for
  Semantic Scholarly Knowledge.
\newblock In \emph{Proceedings of the 10th International Conference on
  Knowledge Capture}, K-CAP '19, 243–246. New York, NY, USA.

\bibitem[{Meister(2017)}]{Meister2017TowardsAK}
Meister, V.~G. 2017.
\newblock Proceedings of the Doctoral Consortium at the 16th International
  Semantic Web Conference ISWC 2017 , Vienna Austria, October 22nd, 2017.
\newblock volume 1962 of \emph{{CEUR} Workshop Proceedings}. CEUR-WS.org.

\bibitem[{Nasar, Jaffry, and Malik(2018)}]{10.1007/s11192-018-2921-5}
Nasar, Z.; Jaffry, S.~W.; and Malik, M.~K. 2018.
\newblock Information Extraction from Scientific Articles: A Survey.
\newblock \emph{Scientometrics}, 117(3): 1931–1990.

\bibitem[{Olson, Zimmerman, and Bos(2008)}]{Gary2008}
Olson, G.~M.; Zimmerman, A.; and Bos, N. 2008.
\newblock \emph{Scientific collaboration on the Internet}.
\newblock The MIT Press.

\bibitem[{Peroni(2014)}]{alma991016811776206926}
Peroni, S. 2014.
\newblock \emph{Semantic Web Technologies and Legal Scholarly Publishing
  [electronic resource] / by Silvio Peroni.}
\newblock Law, Governance and Technology Series, 15.

\bibitem[{Serdyukov, Rode, and Hiemstra(2008)}]{10.1145/1458082.1458232}
Serdyukov, P.; Rode, H.; and Hiemstra, D. 2008.
\newblock Modeling Multi-Step Relevance Propagation for Expert Finding.
\newblock In \emph{Proceedings of the 17th ACM Conference on Information and
  Knowledge Management}, CIKM '08, 1133–1142. New York, NY, USA: Association
  for Computing Machinery.
\newblock ISBN 9781595939913.

\bibitem[{Tang et~al.(2008)Tang, Zhang, Yao, Li, Zhang, and
  Su}]{10.1145/1401890.1402008}
Tang, J.; Zhang, J.; Yao, L.; Li, J.; Zhang, L.; and Su, Z. 2008.
\newblock ArnetMiner: Extraction and Mining of Academic Social Networks.
\newblock In \emph{Proceedings of the 14th ACM SIGKDD International Conference
  on Knowledge Discovery and Data Mining}, KDD '08, 990–998. New York, NY,
  USA: Association for Computing Machinery.
\newblock ISBN 9781605581934.

\bibitem[{Xia et~al.(2016)Xia, Liu, Lee, and Cao}]{Xia_2016}
Xia, F.; Liu, H.; Lee, I.; and Cao, L. 2016.
\newblock Scientific Article Recommendation: Exploiting Common Author Relations
  and Historical Preferences.
\newblock \emph{{IEEE} Transactions on Big Data}, 2(2): 101--112.

\end{thebibliography}
\end{document}